\documentclass[letterpaper, 10 pt, conference]{ieeeconf}  % Comment this line out if you need a4paper
\IEEEoverridecommandlockouts                              % This command is only needed if 
\usepackage{microtype}
\usepackage{comment}

\usepackage{graphicx} % for pdf, bitmapped graphics files
\graphicspath{{./figures/}}
\usepackage{subcaption}
\usepackage{float}
\usepackage{grffile} 

\usepackage{amsmath} % assumes amsmath package installed
\usepackage{amssymb}  % assumes amsmath package installed
\usepackage{units}
\usepackage{bm} %bold italic vectors
\usepackage{multirow}

\usepackage{algorithm}
\usepackage{algpseudocode}

\makeatletter
\let\NAT@parse\undefined
\makeatother

\usepackage[square, numbers, sort&compress]{natbib}
\usepackage{cleveref}

\usepackage[font=small]{caption}
\usepackage{xparse}
\usepackage{xcolor}
\usepackage{tabularx}
\usepackage{diagbox}
\usepackage{makecell}

\newcommand\Tstrut{\rule{0pt}{2.6ex}}         % = `top' strut
\newcommand\Bstrut{\rule[-0.9ex]{0pt}{0pt}}   % = `bottom' strut

\title{\LARGE \bf Sim-to-Real Learning of Footstep-Constrained \\ Bipedal Dynamic Walking}

\author{Helei Duan, Ashish Malik, Jeremy Dao, Aseem Saxena, Kevin Green, Jonah Siekmann$^\dag$, \\Alan Fern, Jonathan Hurst% <-this % stops a space
\thanks{*This work is supported by the NSF Grant No. IIS-1849343, DGE-1314109, and DARPA Contract W911NF-16-1-0002.}% <-this % stops a space
\thanks{All authors are with Collaborative Robotics and Intelligent Systems Institute, Oregon State University, Corvallis, Oregon, 97331, USA. Email: \{{\tt\footnotesize duanh, malikas, daoje, saxenaa, greenkev, siekmanj, afern, jhurst@oregonstate.edu}\}. }
\thanks{$^\dag$Also with Agility Robotics, Albany, Oregon, 97321, USA.}
}

\begin{document}

\maketitle
\thispagestyle{empty}
\pagestyle{empty}

% !TEX root =  main.tex
\begin{abstract}
Recently, work on reinforcement learning (RL) for bipedal robots has successfully learned controllers for a variety of dynamic gaits with robust sim-to-real demonstrations. In order to maintain balance, the learned controllers have full freedom of where to place the feet, resulting in highly robust gaits. In the real world however, the environment will often impose constraints on the feasible footstep locations, typically identified by perception systems. Unfortunately, most demonstrated RL controllers on bipedal robots do not allow for specifying and responding to such constraints. 
This missing control interface greatly limits the real-world application of current RL controllers. 
In this paper, we aim to maintain the robust and dynamic nature of learned gaits while also respecting footstep constraints imposed externally.   
We develop an RL formulation for training dynamic gait controllers that can respond to specified touchdown locations. We then successfully demonstrate simulation and sim-to-real performance on the bipedal robot Cassie. In addition, we use supervised learning to induce a transition model for accurately predicting the next touchdown locations that the controller can achieve given the robot's proprioceptive observations. This model paves the way for integrating the learned controller into a full-order robot locomotion planner that robustly satisfies both balance and environmental constraints. 

\end{abstract}
%%%%%%%%%%%%%%%%%%%%%%%%%%%%%%%%%%%%%%%%%%%%%%%%%%%%%%%%%%%%%%%%%%%%%%%%%%%%%%%%
% !TEX root =  main.tex
\section{Introduction} 

Reinforcement learning (RL) for bipedal robots has shown recent success in learning controllers that can transition between a variety of agile gaits, such as walking, running, jumping, and skipping, with robust sim-to-real transfer \cite{Xie2019, bipedalgaitsICRA, 9636467}. Bipedal animals also exhibit dynamic transitions between various gaits, but in addition have impressive control over specific foot placements that may depart from natural gaits in order to traverse discrete terrains. In contrast, most RL approaches have not yet supported controllers that can account for footstep constraints, but rather allow controllers to choose any footstep locations in order to maintain balance. This significantly limits the applicability of current RL controllers to real-world applications with unsteppable regions, such as potholes, or where precise foot placements are desired.

In this work, we explore RL for bipedal controllers that can both robustly maintain balance while also dynamically satisfying specific footstep constraints that are provided as additional control inputs. Ideally, the learned controllers should satisfy the real-time footstep constraints when they are dynamically feasible, while prioritizing maintaining balance when the constraints are likely to lead to instability or falling. Our approach extends prior work on RL for periodic gaits \cite{bipedalgaitsICRA} to incorporate footstep constraints into the reward function and to train on randomized footstep constraints in simulation. The resulting gait controller is able to both perform periodic gaits by providing constant step length constraints as well as follow sequences of varying step length and direction commands within the robot's capabilities.

\begin{figure}[t]
    \centering
    \includegraphics[width=0.6\columnwidth, scale=1]{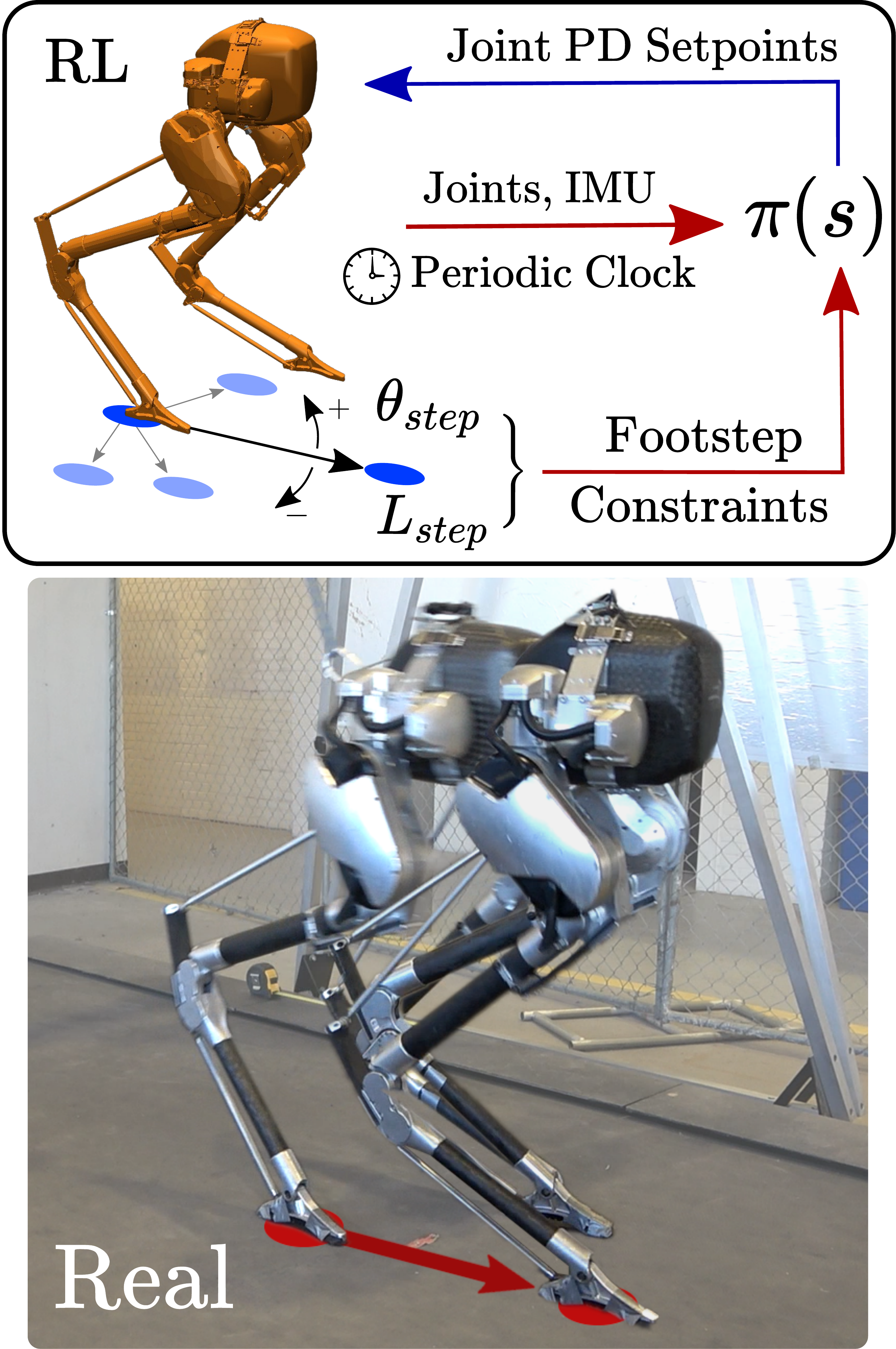}
    \caption{We aim to examine how to maintain the dynamic gaits from learning methods while asking the robot to satisfy specific footstep constraints. The picture shows Cassie performing a controlled 1-2 step maneuver satisfying the desired step command (a forward half meter step) from stepping in place and back to dynamic balancing. 
    % The demonstrated behavior improves from the previous methods by having both dynamic and robust gaits and explicitly control over footsteps. 
    Much of the previous works in RL-based gait policies is incapable of such control authority, rather allowing the full freedom of footsteps.
    }
    \label{fig:cassie_real}
    \vspace{-.1cm}
\end{figure}

We demonstrate this approach on the bipedal robot Cassie in both simulation and on hardware as shown in Figure \ref{fig:cassie_real}. The learned controller shows robustness to varying constraint sequences, external perturbations, and uneven terrain while typically achieving the step targets. In addition, in order to support the use of this controller for higher-level motion planning, we explore supervised learning for building a touchdown to the next touchdown (TD2TD) transition model that is able to predict the reachable set of touchdown locations inferred solely from observable robot states. We show that an accurate TD2TD model can be learned, which sets up future work on a full-body multi-step look-ahead model-predictive planning.
% !TEX root =  main.tex
\section{Related Work}

\textbf{RL for Bipedal Robots with Sim-to-Real.} Learning-based control approaches have demonstrated impressive locomotion on real bipedal robots. A common practice is to reward imitation of a library of template motions \cite{Peng2017, Xie2019, Siekmann-RSS-20, cassieTaskspace, Green2021}. An alternative approach is to design physically-principled rewards \cite{bipedalgaitsICRA} that are coupled with a set of behavior parameters, such as the swing-to-stance ratio of each leg. Both have shown successful demonstrations with sim-to-real. These learned controllers have even successfully traversed flights of stairs with no perception \cite{Siekmann-RSS-21}.
To allow practical use of learned solutions on bipedal robots, learned policies should have the ability to respect footstep constraints that come from the external environment, which theoretically will provide more control over the locomotion behaviors and thus will enable the robot to perform tasks under the constraints of the real world.

\textbf{Footstep Constrained Learning for Legged Robots.} Legged robots have a unique advantage in that they can dynamically choose sparse footholds to overcome challenging terrains. Several previous studies focused on using learning methods to allow robots to reach specific target foot touchdown locations. 
Some learning-based approaches often use known models or heuristics to assign the target touchdown locations before any locomotion behaviors are learned \cite{deepgait, DBLP:conf/corl/DaXHBAZBG20}, while others train gait policies that can maximally track the target footsteps \cite{Peng2017, Xie2020a}. In particular, our work is inspired by \citet{Xie2020a}, where the authors bootstrapped a pretrained locomotion policy with curriculum learning to follow target footsteps in simulation. By contrast, our approach trains a gait policy from scratch and does not use a curriculum to maximize performance. Despite this, we realize a successful sim-to-real transfer, potentially due to the physics-principled periodic reward design.   

\textbf{Transition Model Learning for Legged Robots.} 
Study of bipedal locomotion heavily relies on parameterizing gaits into simplified representations along with reduced-order dynamics models. Many of them define control actions based on gait events, such as touchdown, toe-off or apex height during aerial phase. These compact models \cite{Geyer2006, Wu2013, Kajita2001} are widely used in control and planning frameworks on bipedal robots to decide next footsteps or future position and force trajectories \cite{Martin2017a, Apgar2018a, 9561961}. While they can capture core underactuated dynamics of locomotion, there still exists a significant gap between reduced-order models and full-order robots, such as the mismatch of kinematics and inertia properties. Previous studies have focused on improving these models with an additional learning process on top of reduced-order models \cite{9562093, 9562022, DBLP:journals/corr/abs-2104-09771}. Another approach is to build up or refine the transition model from the bottom-up, by sourcing the data from some working controllers \cite{Coros2008, Bledt2020, Morimoto2007a}. Given the full-order robot control policies, we seek to use neural networks to learn a mapping from proprioceptive robot states to predict the reachability for next touchdown locations. As a result, planning will become viable with this learned transition model. 

% !TEX root =  main.tex

\section{Learning Footstep-Constrained Gait Policy}
\label{sec:gait policy}
\noindent

Our formulation of learning with footstep constraints builds on and extends the RL framework from prior work on bipedal locomotion \cite{bipedalgaitsICRA}. As outlined below, a key difference from this prior work is that our control policy receives command inputs that constrain relative footsteps (see Figure \ref{fig:method_annotation}), rather than commands that constrain velocities and periodic gait characteristics. Accordingly, this difference requires adapting the reward formulation to encourage matching footstep commands rather than periodic gait properties. 

\subsection{State Space}
The state space of the policy consists of the observed robot states and user commands with a size of $\mathbf{S \in \mathbb{R}}^{42}$.
% The proprioceptive sensing inputs to the neural network are measured through the proprietary state-estimator from Agility robotics. 
More specifically, $\mathbf{S}$ includes the base's rotational velocities ($\mathbb{R}^{3}$), orientations in quaternion ($\mathbb{R}^{4}$), and actuated and un-actuated joint positions and velocities measurable on hardware ($\mathbb{R}^{28}$). 
$\mathbf S$ also includes a 2-dimensional periodic clock input that is calculated using two offset sine functions to create unique phase identifiers and wrap around the cycle (see prior work \cite{bipedalgaitsICRA} for clock details). The state also includes the command inputs representing the desired swing-to-stance ratio ($\mathbb{R}^{2}$) and a turn rate input ($\mathbb{R}^{1}$) that allows the robot to offset the absolute angles for facing forward. The final command input component of the state specifies the desired footstep constraint $U$. Specifically, this command specifies the relative offset in polar coordinates ($\mathbb{R}^2$) between the current stance foot location and the desired next TD location of the same foot, defined as $U=(L_{step},\theta_{step})$. 

\subsection{Action Space}
The action space $\mathbf{A \in \mathbb{R}}^{10}$ is defined as the PD setpoints for each motor. The action is updated at \unit[40]{Hz} with the policy. The underlying joint PD controller receives the policy actions and runs fixed-gain joint PD control at \unit[2]{kHz}.

\subsection{Reward}

In order to train the robot to both maintain balanced locomotion and match footstep constraints we use rewards at two different time scales. The reward for balanced locomotion is a dense reward that occurs at every decision cycle, while the footstep-constraint reward is sparse and occurs only at TD moments.  

The primary goal for the dense reward is to encourage the swing and stance behaviors for each foot as defined by the clock. In addition, the dense reward asks the robot to move towards the new footstep targets by checking the distance traveled between policy updates. This term is computed as $r_{pelvis}=100(\Delta d^i - \Delta d^{i-1})$, where $d^i$ is the distance from the target to the pelvis at $i$th policy step. If the step distance is smaller then 0.1m, this term alternatively encourages stable pelvis motions by penalizing large velocities $r_{pelvis}=1-(\exp(-3|\dot{x}|) + \exp(-3|\dot{y}|))$. 

At every TD event indicated by the clock input, the sparse reward is computed from the Euclidean distance $f$ between the actual and target footstep locations as $r_{step}=\exp(-2f)$. It will be added on top of the dense reward at every footstep. Although the sparse bonus could become a dense one by adding it throughout the entire swing phase, we found this will lead to aggressive and physically infeasible swing behaviors. The remaining reward terms listed in Table \ref{table:reward_description}, such as orientation and action smoothness, are defined the same as in \cite{bipedalgaitsICRA}, which we refer readers to for detailed reward definitions.
\begin{table}[!h]
% \vspace{-.1cm}
\centering
\resizebox{1\columnwidth}{!}{%
\begin{tabular}{l|l||l|l}
\hline
Name & Weight & Name & Weight \\ \hline
 Right foot force & 0.155 & Left foot force & 0.155 \\ \hline
 Right foot speed & 0.125 & Left foot speed & 0.125 \\ \hline
 Robot orientation & 0.125 & Pelvis stability & 0.125 \\ \hline
 Action smoothness & 0.0325 & Torque smoothness & 0.0325 \\ \hline
\textbf{Pelvis} & \textbf{0.125} & \textbf{Sparse step reward} & \textbf{3.0} \\ \hline
\end{tabular}}
\caption{Reward Terms and Weights.}
\label{table:reward_description}
% \vspace{-.7cm}
\end{table}

\subsection{Policy Commands Generation}
\label{sec:command_generation}
\begin{figure}[t]
    \centering
    \includegraphics[width=\columnwidth]{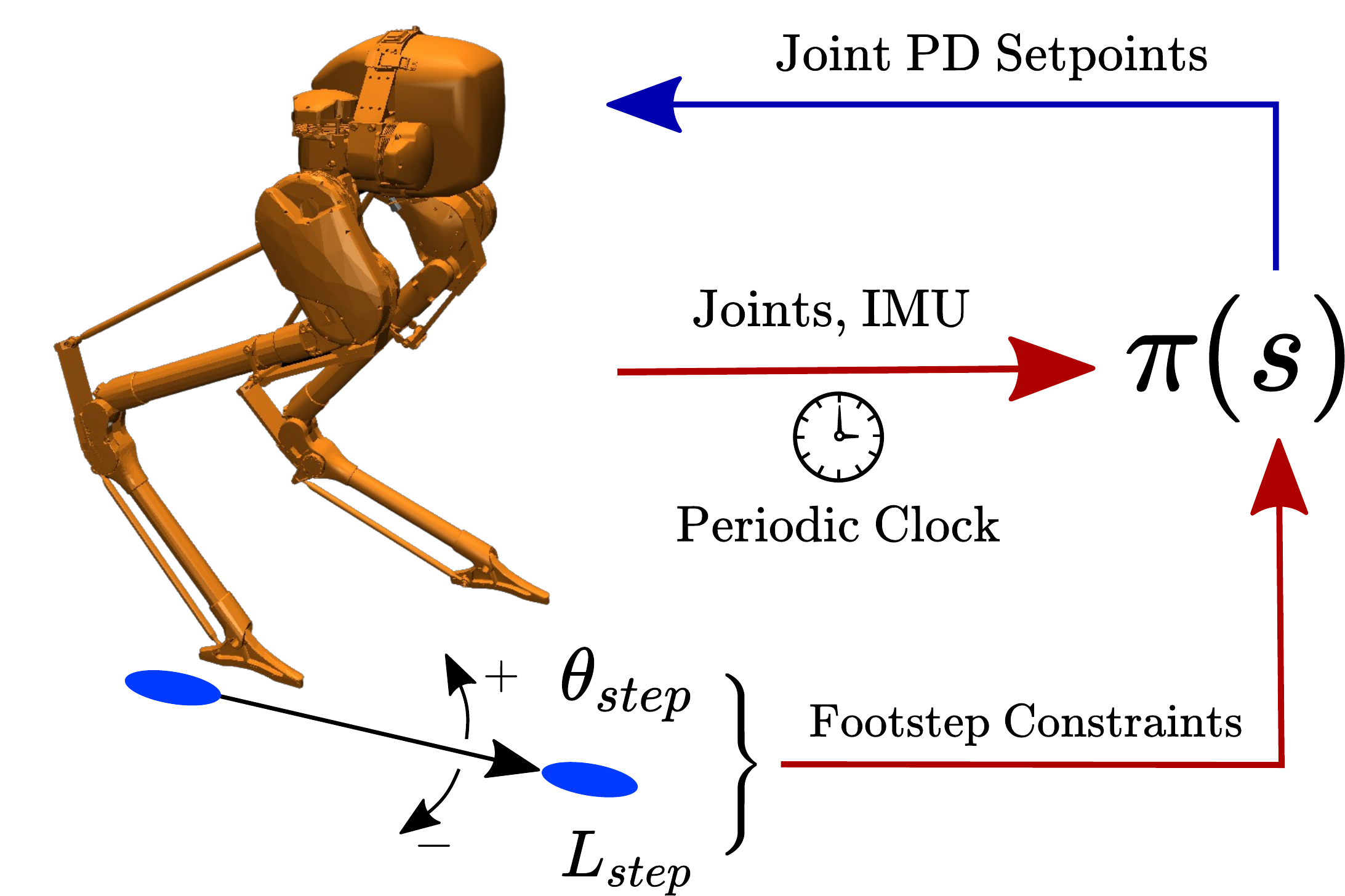}
    \caption{The footstep-constrained gait policy maps proprioceptive information of joints and IMU sensors into joint PD setpoints. Importantly, the control policy is conditioned on footstep commands that are relative to the previous touchdown location for the same foot. The commands are updated at a random TD event and parameterized in the polar coordinate $(L_{step}, \theta_{step})$. Positive $\theta_{step}$ means the target touchdown location is left from the previous TD and vice versa. 
    Although the policy uses polar coordinate for commands, the data representations are all plotted in square for convenience. 
    }
    \label{fig:method_annotation}
\end{figure}

The nature of the reward structure introduced in \cite{bipedalgaitsICRA} provides a clock for timing the swing and stance phases of the gait. 
When the robot learns how to swing the leg, it could potentially explore swinging towards the desired step target and would be rewarded for doing so. 
We use the clock values for each leg to infer the touchdown event by checking against a threshold value. 
Once the clock value passes the threshold and indicates a touchdown, we generate a new footstep command relative to the current stance foot. We do not use forces to check TD events in order to generate the new command or compute sparse reward to avoid explicit force sensing on hardware. 
% \yesh{The footstep target used in the reward function is computed in global coordinates by adding the generated command, as shown in Figure \ref{fig:method_annotation}.} 
The footstep target used for the reward function is subsequently computed in global coordinates by adding the command with the current stance foot location as shown in Figure \ref{fig:method_annotation}. 
The policy input is also updated with the generated command, expressed as an offset from the last commanded touchdown position in polar coordinates.
As a result, the policy has a one-step preview for the same side footstep target.

Each roll-out will be terminated if the robot falls by checking for large orientation changes. We do not use hard termination conditions when the robot fails to reach the target for a particular step. In other words, the robot can keep collecting rewards even if the touchdown locations are far from the desired ones. 
Since the footstep target generation is aligned with the periodic clock, the policy is encouraged to sync up its gait with the swing footstep targets in order to maximize the reward.

From biomechanics observations, bipedal locomotion can be viewed as a constrained optimization problem among traveling speed, step frequency, and step length \cite{Bertram2005}. Here we do not focus on tracking any commanded speed, but rather emphasize the step frequency and commanded footstep targets. Thus, step frequency becomes a critical control parameter and majorly affects the swing timing. The swing time influences how the robot can reach towards the next footstep target within the actuator's capability. 

The periodic clocks cover a range of swing-to-stance ratios and step frequencies. The clock is represented with a cyclic phase variable $\phi$, and is computed as, 
$
\phi^{i+1} = \phi^{i} + \gamma \Delta \phi^{i}
$
, where $i$ is the policy update step, $\Delta \phi$ is the nominal increment, and $\gamma$ is the phase multiplier that controls the scale of the phase increment, resulting in different step frequency. 

Instead of having a constant $\gamma$, we randomize different combinations of ratio and step frequency, asking the policy to try to gather higher rewards by exploring the dynamics as much as possible. During each roll-out, to allow for moderate randomization, there is a 1/50 chance of randomizing the control parameters (swing ratio, step frequency, and footstep commands) at each policy step. Note that the footstep commands may be randomized at every policy step, but are only updated per TD event. 
All control parameters are randomized with uniform sampling from a set of ranges shown in Table \ref{table:control_randomization}, 

\begin{table}[!h]
% \vspace{-.2cm}
\centering
\resizebox{0.75\columnwidth}{!}{%
\begin{tabular}{l|l|l}
\hline
Control Parameter & Unit & Range \\ \hline
 Swing-to-Stance Ratio                      & --   &   $[0.45, 0.6]$ \\ \hline
 Step Frequency $\gamma$                         & --        &   $[0.9, 1.3]$     \\ \hline
 Footstep Length    $L_{step}$            & m   &   $[0, 0.8] $     \\ \hline
 Footstep Direction $\theta_{step}$               & rad  &   $[-\pi, \pi]$     \\ \hline
\end{tabular}}
\caption{Control parameters are randomized during training.}
\label{table:control_randomization}
% \vspace{-.2cm}
\end{table}

\subsection{Learning Algorithm and Setup}
We use the Proximal Policy Optimization (PPO) algorithm \cite{Schulman2017}, a model-free policy gradient method with an actor-critic network. Both the actor and critic use LSTM neural networks with two recurrent hidden layers of size 128. Actor and critic networks are independent of each other and do not share weights between them. We used the clipped objective for PPO based on the KL divergence between the updated policy and the previous policy. Following \cite{Siekmann-RSS-20}, we also deploy dynamics randomization during training. 
% !TEX root =  main.tex

\newpage
\section{Footstep Behavior Prediction with Learned Transition Model}
\label{sec:model learning}

With the learned footstep-constrained gait policy, we aim to understand how predictable the next touchdown is by introducing a predictive model. This model tries to capture the TD2TD transition dynamics assuming the robot is controlled under the policy. Specifically, this model maps the robot states into the predicted step error, the Euclidean difference between the desired and actual footstep locations. This model spans the 2D command space, quantifying the step commands from the current state. 

\subsection{Data Collection}
We begin with data collection given a trained footstep policy. The intention is to capture the 1-step behavior right after the robot takes a new step command. First, the 2D step command space is discretized into a 30$\times$30 grid in terms of Cartesian distance from the current stance foot location. We convert the Cartesian distance back to polar commands for policy inputs. In the beginning of each collection episode, we reset the simulation and policy into a random state, including a randomized step command. The policy then executes for an initial number of touchdowns to settle the robot into a steady-state gait. At the 5th TD event on either the left or right foot, we record the robot state and then change to a new command. Finally, we record the subsequent touchdown with step error for that new command. This process is repeated for each step command in the 30$\times$30 grid using the same reset state. To effectively cover a wide range of the state-command space, we can increase the data coverage by having many different reset states.

\subsection{Model Definition \& Training}
The model uses high-dimensional robot states to predict the step error for a range of step commands as shown in Figure \ref{fig:prediction}. We use a neural network to map between these high-dimensional spaces. Specifically, the input $\mathbf{X \in \mathbb{R}}^{42}$ is defined the same as the policy input $\mathbf{S}$. The output $\mathbf{Y \in \mathbb{R}}^{30\times30}$ represents a range of step errors spanning the command space. Each value of the output is the predicted step error from the current stance foot measured in the Cartesian distance.

One characteristics of the output is the continuity among the nearby step commands, because step commands that are next to each other can result in approximately similar touchdown behaviors in terms of step errors. Thus, we use an CNN as the network structure by applying multiple layers of 2D transposed convolution operators over the robot state. To train the model, we apply the mean-square loss over the entire grid, standard train-test split, and batch training.
% !TEX root =  main.tex
\section{Results}

We use \emph{cassie-mujoco-sim} simulation package\footnote{https://github.com/osudrl/cassie-mujoco-sim} based on the Mujoco physics simulator \cite{Todorov2012} for the RL sampling process. Each roll-out has 300 simulation timesteps, equivalent to approximately 7.5 seconds with roughly 10 touchdowns for each foot. We empty the replay buffer at the beginning of every iteration and sample 50,000 new simulation samples before doing optimization. For optimization, each batch consists of 64 rollouts. For both the actor and critic, we use the Adam optimizer with a learning rate of $0.0005$ for 5 epochs per iteration. The optimization is terminated if the KL divergence is greater than 0.02. The policy was trained for approximately 400 millions timesteps, using 56 cores on a dual Intel Xeon Platinum 8280 server on Intel vLab. 

% We now describe results of the learned footstep-constrained policy in simulation and on the robot Cassie. 
The learned policy is first evaluated under constant footstep constraints. Then it is tested with randomized footstep commands. We show the learned dynamic gaits with sim-to-real transfer under constant and one-off footstep constraints on the bipedal robot Cassie. Lastly, we present the predicted reachability map from the learned 1-step transition model. 

\subsection{Simulation Results}

\begin{figure}[t]
    \centering
    \includegraphics[width=0.9\columnwidth]{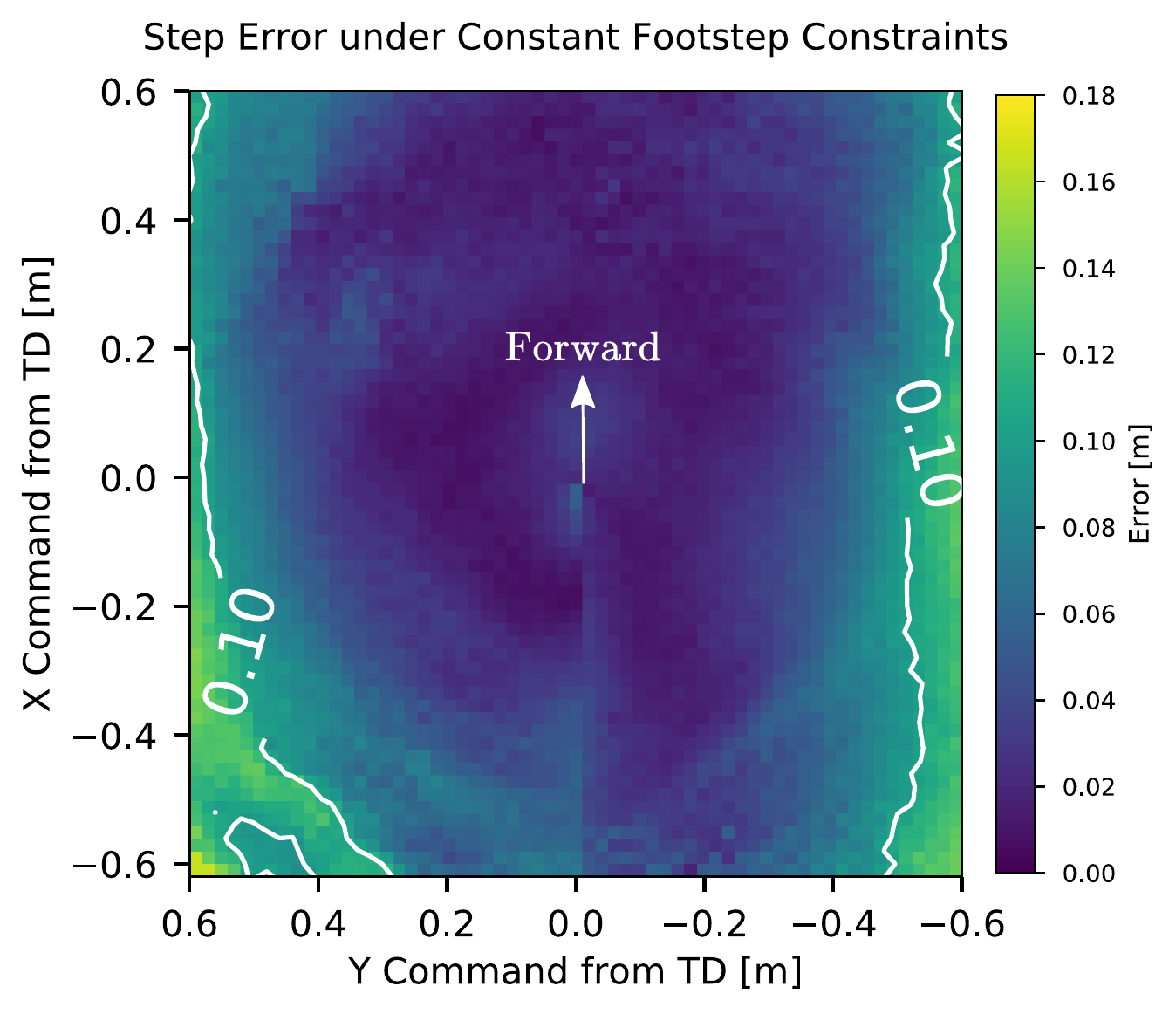}
    \caption{Each pixel of the image shows the step error for the corresponding next same side footstep command. Commands are for the same foot that is currently in touchdown and is in relative X and Y distance away from the current touchdown location. For example, X command of \unit[0.6]{m} and Y command of \unit[0]{m} represents walking straight ahead, each foot having relative step length of \unit[0.6]{m} and direction of 0$^{\circ}$ from the previous same side touchdown location. Step error is the Euclidean distance between the target and actual touchdown locations. We see that large side (Y command greater than 0.5m) and diagonal steps tend to have larger errors. Interestingly, due to the polar command definition, we notice the circular patterns of the contour levels. 
    % Also, due to the wrap around of the polar command crossing $-\pi$ to $\pi$, we observe a visible discontinuous change but not reducing the performance. 
    }
    \label{fig:constant_cmds}
\end{figure}

We perform various experiments in simulation to validate and quantify the behaviors of the footstep policy. First, we evaluate the step accuracy when the policy is controlling Cassie with constant footstep constraints. This will mimic speed control, but also satisfies the footstep constraint every step. We sweep through the step command space within the range of step length and direction. For each step command, we record the step errors for 20 consecutive steps. As shown in Figure \ref{fig:constant_cmds}, the majority of the commands have errors less then 0.1m. The step error tends to increase with large commands. Cassie's narrow feet have no control over frontal plane motions which means large side and diagonal steps while the pelvis is facing forward are more challenging.

The second experiment intends to capture the variability needed between consecutive step commands, meaning the next footstep often has a small change from the previous one. To test the footstep behaviors under random footstep constraints, we define four categories of tests with increasing levels of difficulty. Each category has different maximum command changes from the previous footstep command. The command changes are uniformly sampled within a limit, and then added on top to the previous step command. 
We reject commands less than 0.01m to prevent them stuck within small footsteps. 
For each test, we collect 200 footsteps in simulation and track the step errors.

Additionally, we investigate briefly on the effects of step frequency onto the step accuracy in simulation. It has been previously shown that adapting the step frequency can greatly robustify locomotion \cite{Nishiwaki2010StrategiesFA, 7803251}. We experiment this by changing the phase multiplier $\gamma$ explained in Section \ref{sec:command_generation}.
When we increase the footstep command, we would like to increase swing time (decrease $\gamma$) so the policy has more time to move the foot toward the target. 
% In the other direction, when we decrease the footstep command we would like to decrease swing time (increase $\gamma$) to encourage the policy to put the foot down sooner.

We use three different setups to test this effect. First setup (Fixed $\gamma$) is the base policy that is trained with a range of swing-to-stance ratios and step frequencies. 
But for this experiment, the base policy is evaluated with the nominal ratio (0.45) and step frequency ($\gamma=1$). 

Second setup (Linear $\gamma$) still uses the base policy, but changes the step frequency in a linear relationship with the command footstep length $L_{step}$, expressed as a remapping,
\[
\gamma=\bigg(1 - \frac{L_{step}}{L^{max}_{step}}\bigg)(\gamma^{max}-\gamma^{min}) + \gamma^{min}.
\]
Third setup (Heuristic $\gamma$) is a policy that is trained with a step frequency heuristics to regulate $\gamma$. Linear $\gamma$ uses the one time information about the step command to vary $\gamma$. In this setup, we seek to maintain a history of information on the step command to vary $\gamma$. 
We keep a weighted moving average of the step commands $U^{avg} = (L^{avg}, \theta^{avg})$. Then we use the difference between the target $\Delta = ||U^{tar} - U^{avg}||_2$ to modify the phase multiple $\gamma$ as the following: 
\begin{equation}
  \gamma =
  \begin{cases}
    1.0 & \text{if } \Delta < 0.2, \\
    \Big(\frac{L^{avg}}{L^{tar}}\Big)^2 & \text{if } \Delta \geq 0.2 \text{ and } L^{avg} \geq L^{tar} \text{ and } \\
    & \hspace{6.5em} |\theta^{avg}-\theta^{tar}| \leq \frac{\pi}{10}, \\
    \frac{1}{(0.8 + \Delta)^2} & \text{otherwise}
  \end{cases}
  \nonumber
\end{equation}
For small changes in command, there is no need to change $\gamma$ so we keep it at 1. 
In the second case, we want to make $\gamma$ greater than 1 to decrease swing time and make touchdown happen sooner. Otherwise, we make $\gamma$ less than 1 to increase swing time and give the foot more time to reach the target.

From Table \ref{tab:sim_random_cmds}, we see that step errors become larger when the difficulty of commands increases. When the step length is the same, randomizing the step direction causes the performance to drop. In general, having step frequency modifications (Linear and Heuristic $\gamma$) can improve the step accuracy. However, the hand-designed heuristic is only based on the footstep commands during training. The step frequency might also be dependent on other dynamics \cite{Bertram2005}, such as energetic cost, body velocities, or momentum. We leave the smarter step frequency regulation as future work. 

\begin{table}
\centering
\vspace{.3cm}
\resizebox{0.45\textwidth}{!}{%
\begin{tabularx}{\columnwidth}{c|X|X|X|X}
\hline
% \backslashbox{$\gamma$ Type}{$\Delta U$}
\backslashbox[17mm]{$\gamma$ Type}{Range of\\$\Delta U$}
& \thead{$\pm0.3m$,\\$\pm0^{\circ}$} & \thead{$\pm0.3m$,\\$\pm20^{\circ}$} & \thead{$\pm0.7m$,\\$\pm0^{\circ}$} & \thead{$\pm0.7m$,\\ $\pm20^{\circ}$} \\ \hline
Fixed\Tstrut\Bstrut  & 0.11$\pm$0.06m\Tstrut\Bstrut    & 0.14$\pm$0.08m\Tstrut\Bstrut     & 0.19$\pm$0.12m\Tstrut\Bstrut    & 0.23$\pm$0.13m\Tstrut\Bstrut     \\ \hline
Linear\Tstrut\Bstrut  & 0.12$\pm$0.06m\Tstrut\Bstrut    & 0.15$\pm$0.09m\Tstrut\Bstrut     & 0.18$\pm$0.13m\Tstrut\Bstrut    & 0.22$\pm$0.14m\Tstrut\Bstrut     \\ \hline
Heuristic\Tstrut\Bstrut & 0.09$\pm$0.05m\Tstrut\Bstrut &  0.14$\pm$0.10m\Tstrut\Bstrut & 0.17$\pm$0.09m\Tstrut\Bstrut & 0.17$\pm$0.10m\Tstrut\Bstrut \\ \hline

\end{tabularx}%
}
\caption{
This table shows the responsiveness of randomizing footstep constraints with mean and std of the step errors.
The next footstep constraints are randomized in an incremental change $\Delta U$ from the last footstep command, expressed as $U^{next}=U^{prev}+\Delta U$.
% Mimicking the real-world case, where the steps are always gradual changes from the last ones. 
By actively changing step frequency, Linear and Heuristic setup can both achieve better accuracy than constant step frequency.}
\label{tab:sim_random_cmds}
% \vspace{-.5cm}
\end{table}

\subsection{Sim-to-Real on Cassie}

The learned policies were transferred onto a bipedal robot Cassie and successfully completed multiple experiments. We used the learned policy with fixed nominal ratio and step frequency to perform the tests. First, we tested the policy under constant footstep constraints. We used the estimated feet positions and pelvis positions to calculate the relative step length of each foot on the hardware. The position data was collected using the proprietary state estimator from Agility Robotics. From Table \ref{tab:hardware_constant}, each footstep constraint is well satisfied except for larger step lengths. We noticed lateral drifting issues when the robot tried to satisfy the large step commands, causing the step length to be shorter then the desired one. 

\begin{table}[H]
\vspace{-.1cm}
\centering
\resizebox{\columnwidth}{!}{%
\begin{tabular}{llll}
\hline
Target [m] & 0.3 & 0.5 & 0.8          \\
\hline
Measured [m] & $0.32\pm 0.03$ & $0.46\pm 0.04$ & $0.62\pm 0.06$ \\
\hline
\end{tabular}%
}
\caption{Step command error of Cassie walking on treadmill with constant footstep constraints. Step errors are measured as the mean and the standard deviation over 50 footsteps. }
\label{tab:hardware_constant}
\vspace{-.6cm}
\end{table}

\begin{figure}[b]
    \vspace{.3cm}
    \centering
    \includegraphics[width=0.9\columnwidth]{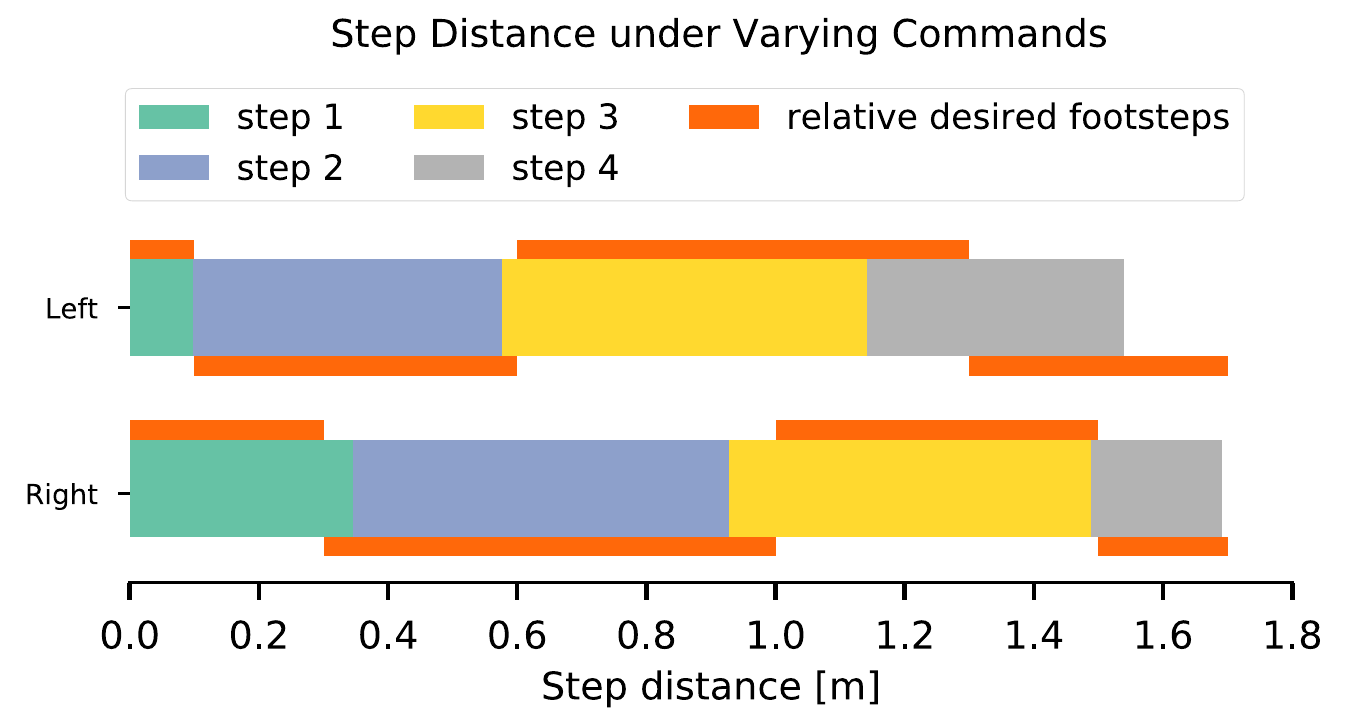}
    \caption{On hardware, Cassie starts with stepping in place and executes a sequence of step commands for each foot, and finally goes back to stepping in place. Similar to Table \ref{tab:hardware_constant}, we noticed undershooting when the commands are larger, which leads to the left foot traveling a shorter distance overall. 
    }
    \label{fig:varying}
\end{figure}

Next, we tested the policy with a sequence of varying step commands. For example, when the robot is walking over gaps with different sizes, the policy should be able to regulate the footsteps. The step commands are lists of step lengths, [0.1, 0.5, 0.7, 0.4]m for the left and [0.3, 0.7, 0.5, 0.2]m for the right. Cassie starts with the left foot as shown in Figure \ref{fig:varying}.

\begin{figure*}[t]
    \vspace{.2cm}
    \centering
    \includegraphics[width=0.93\textwidth]{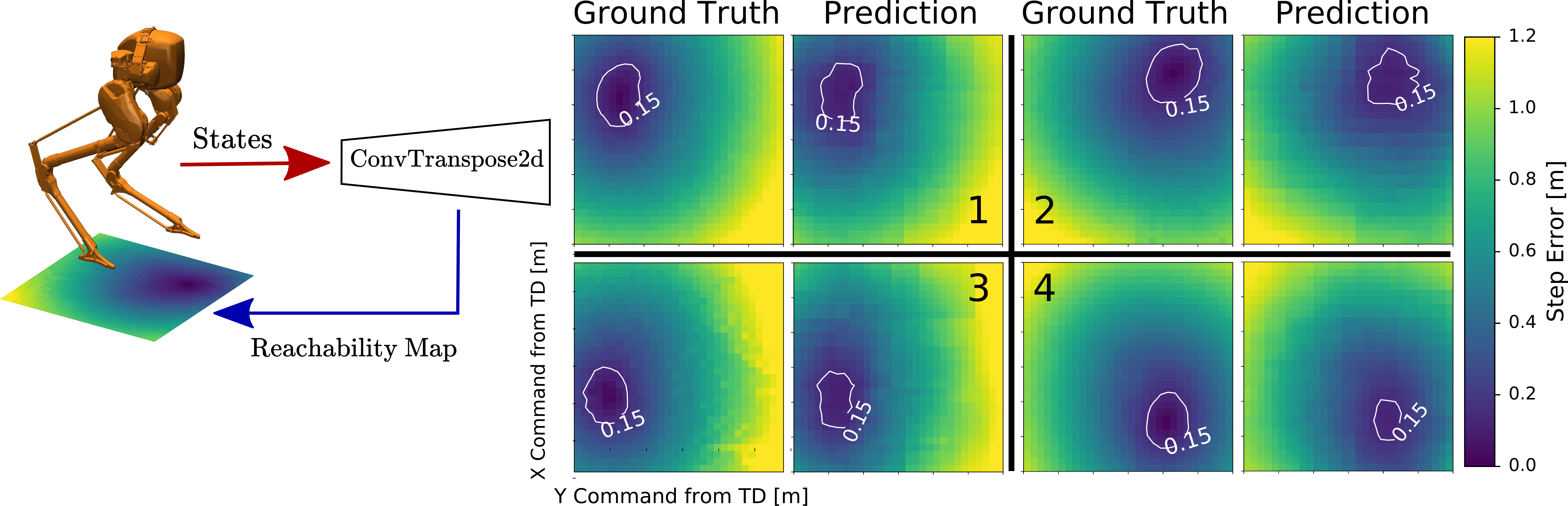}
    \caption{1-Step reachability map is predicted and compared with the ground truth. Four examples show that the model can accurately predict the next footstep behaviors under various robot TD states. Each case has a set of reachable footsteps. Regions within the contour lines are defined as the size of the reachable set, also plotted in Figure \ref{fig:box}. The predicted images are smoother than ground truth in some cases (image 3), due to the image-based techniques.}
    \label{fig:prediction}
    \vspace{-.6cm}
\end{figure*}

Third, Cassie also performed basic 1-2 steps, where Cassie starts stepping in place, takes a desired step, and then goes back to stepping in place. We tested various of 1-2 steps maneuvers on hardware with varying directions and ground height. Cassie is able to show robust behaviors on unseen height changes when reaching to the step target. Especially when the ground becomes very uneven, the control policy will prioritize the robustness for balance over tracking footsteps. We refer readers to the accompanying video for better visual demonstrations.  

\subsection{Learned 1-Step Transition Model}

\begin{figure}[h]
    \centering
    \includegraphics[width=0.85\columnwidth]{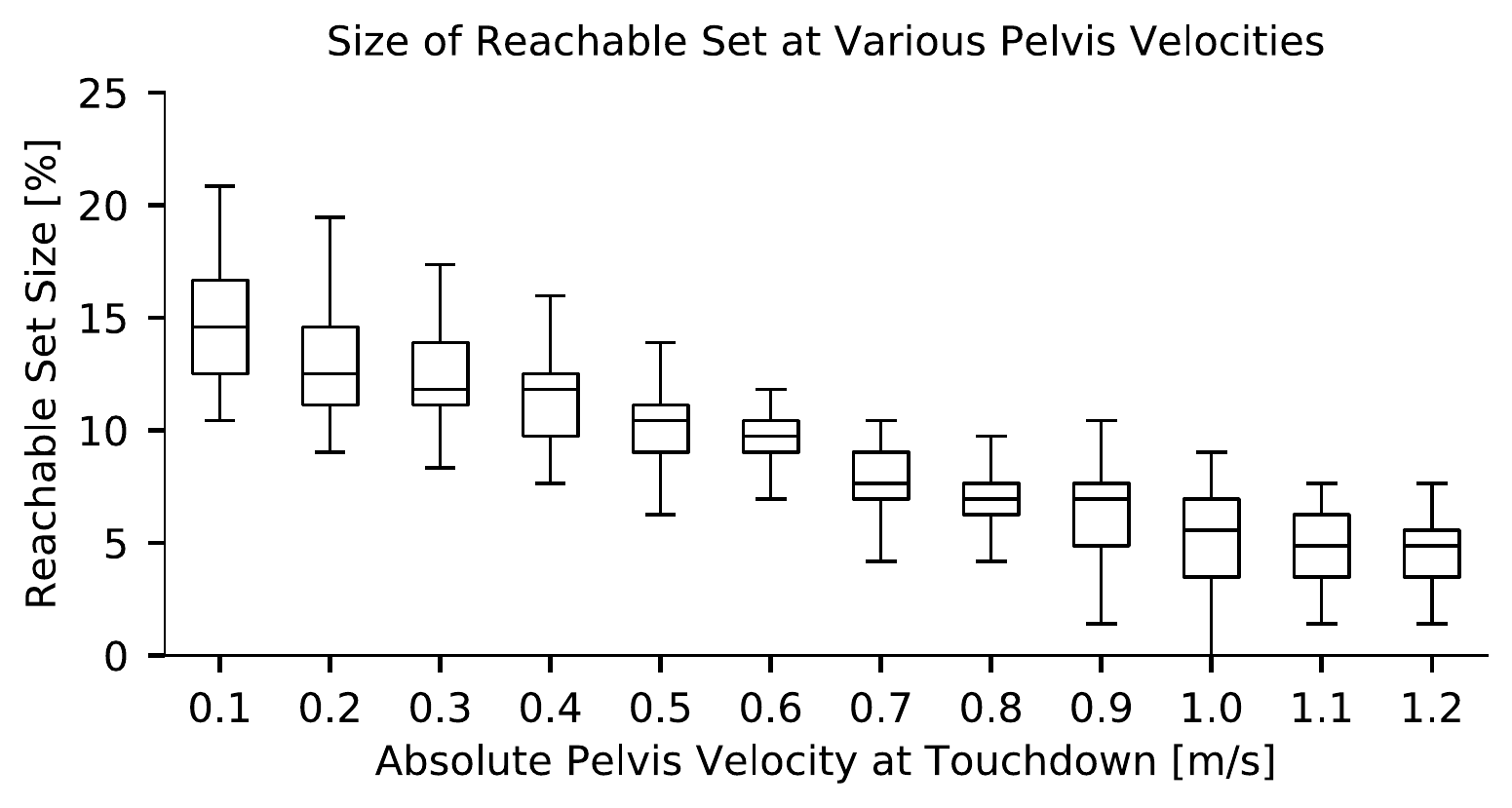}
    \caption{From the collected data, when the robot travels at fast speed regardless of directions, the size of the reachable set (show as median and interquartile-range) becomes smaller, indicating the robot is more committed into a narrower range of commands under the learned policy. The velocities are binned into categories. 
    % We define the reachable sets by selecting step error less than 0.15m for each reset state. 
    }
    \label{fig:box}
\end{figure}

We collected $9\times 10^6$ labeled data from simulation using $10^4$ reset states. We rarely found failure cases leading to the robot falling down, due to the learned policy prioritizing balance first, even though the commands can be vastly different from the previous ones. We are interested to see the size of the reachable set out of the entire command space, defined when the step errors less than 0.15m. Figure \ref{fig:box} shows that the policy is capable of reaching a larger range of next footsteps when the robot velocity is lower. Intuitively, when the robot travels at higher speeds, it will commit more into a narrower set of reachable TD locations, thus needing multi-step planning to resolve the transitions. Figure \ref{fig:prediction} shows four examples of the reachability map predictions under various robot states. We noticed the inductive bias of smooth variation in the output of the proposed network is potentially suitable to this application.

% !TEX root =  paper.tex
\section{Conclusion} 

In this paper, we present a learning approach for bipedal locomotion control under footstep-constraints. Our approach trains the control policy from scratch with two objectives in different timescales, periodic locomotion and sparse footstep-constraints. The control policy is conditioned on the footstep commands. The resulting learned solution shows dynamic gaits while satisfying footstep constraints imposed through user commands. We demonstrate the learned behaviors with various experiments in simulation as well as successful sim-to-real transfer on the robot Cassie.

Our proposed approach can potentially lead to several future research directions by addressing and extending the following aspects. First, we observed that the robot tends to drift when inputting large steady-state step commands. We hypothesize that this is due to error in internal estimation of feet locations. We could alleviate this by directly inputting the feet positions from the estimator. 
Second, although we showed that the learned policy is robust to changes in step height to some extent, actively handling step height could improve previous work on blind stair walking, which treated step height as pure disturbances \cite{Siekmann-RSS-21}. Third, we hope to extend the 1-step prediction model to interact with the environment by learning the reachability model for 2 to 3 steps in the future, which has been shown to be enough look ahead for planning \cite{7803251}. Encoding the full-order robot state into a multi-step reachability prediction can greatly simplify the planning problem with the robust learned control policy. With the ability to actively control 3D footsteps and robustly maintain balance, a reachability map provides a common language to interact with nearby terrains and plan for multi-step behaviors, leading to reliable and robust control capable of walking up and down stairs and avoiding unsteppable regions.

% \input{appendix}
% \addtolength{\textheight}{-8cm}   % This command serves to balance the column lengths

%%%%%%%%%%%%%%%%%%%%%%%%%%%%%%%%%%%%%%%%%%%%%%%%%%%%%%%%%%%%%%%%%%%%%%%%%%%%%%%%

\section*{Acknowledgments}
\small{We thank Intel for providing vLab resources and students at Dynamic Robotics Laboratory for helpful discussions.}

\clearpage
\def\bibfont{\footnotesize}
\bibliographystyle{IEEEtranN}
\bibliography{main}

\end{document}